\pdfoutput=1

\documentclass[11pt]{article}

\usepackage{acl}

\usepackage{times}
\usepackage{latexsym}

\usepackage[T1]{fontenc}

\usepackage[utf8]{inputenc}

\usepackage{microtype}

\usepackage{inconsolata}
\usepackage{import}
\usepackage{microtype}
\usepackage{layout}
\usepackage{tabularx, makecell}
\usepackage{booktabs}
\usepackage{mathrsfs}
\usepackage{amssymb} 
\usepackage{url}
\usepackage{graphicx}
\usepackage{xspace,paralist}
\usepackage{times,latexsym}
\usepackage{amsmath}
\usepackage{appendix}
\usepackage{comment} 
\usepackage{enumitem}
\usepackage{makecell}
\usepackage{multirow}
\usepackage{xcolor}
\usepackage{graphicx}
\usepackage{cleveref}
\usepackage{tcolorbox}
\usepackage{adjustbox}
\usepackage{relsize}
\usepackage{todonotes}
\usepackage{listings}
\usepackage[frozencache,cachedir=minted-cache]{minted} 
\usepackage{xcolor}
\definecolor{bg}{rgb}{0.1,0.1,0.1}
\definecolor{fg}{rgb}{0.9,0.9,0.9}

\setminted{
    bgcolor=bg,
    fontsize=\footnotesize,
    breaklines
}

\newcommand{\tabref}[2][]{Table#1~\ref{#2}\xspace}
\newcommand{\figref}[1]{Figure~\ref{#1}\xspace}

\newcommand{\appref}[1]{Appendix~\ref{#1}\xspace}

\newcommand{\errortype}[1]{\textit{#1}\xspace}
\newcommand{\ftype}{\errortype{Type1}}
\newcommand{\stype}{\errortype{Type2}}
\newcommand{\ttype}{\errortype{Type3}}

\newcommand{\model}[1]{\text{#1}\xspace}

\newcommand{\chatgpt}{\model{ChatGPT}}
\newcommand{\gptthreepointfive}{\model{GPT-3.5-Turbo}}
\newcommand{\gptfour}{\model{GPT-4}}

\newcommand{\llama}{\model{LLaMA}}
\newcommand{\llamatwo}{\model{LLaMA-2}}

\newcommand{\davinci}{\model{Davinci-text-003}}

\newcommand{\pplai}{\model{PerplexityAI}}

\newcommand{\checker}[1]{\textit{#1}\xspace}
\newcommand{\rarr}{\checker{RARR}}
\newcommand{\factscore}{\checker{FactScore}}
\newcommand{\factool}{\checker{FacTool}}
\newcommand{\factcheckgpt}{\checker{Factcheck-GPT}}
\newcommand{\cove}{\checker{CoVe}}
\newcommand{\fire}{\checker{FIRE}}
\newcommand{\perplexityai}{\model{Perplexity.ai}}

\newcommand{\dataset}[1]{\text{#1}\xspace}
\newcommand{\factqa}{\dataset{FactQA}}
\newcommand{\snowball}{\dataset{Snowball}}
\newcommand{\freshqa}{\dataset{FreshQA}}
\newcommand{\selfaware}{\dataset{SelfAware}}
\newcommand{\factoolqa}{\dataset{FacTool-QA}}
\newcommand{\felmwk}{\dataset{FELM-WK}}
\newcommand{\factcheckbench}{\dataset{Factcheck-Bench}}
\newcommand{\factscorebio}{\dataset{FactScore-Bio}}
\newcommand{\halueval}{\dataset{HaluEval}}
\newcommand{\factbench}{\dataset{FactBench}}

\newcommand{\module}[1]{\textsc{#1}\xspace}
\newcommand{\ofc}{\model{OpenFactCheck}}
\newcommand{\custchecker}{\module{CustChecker}}
\newcommand{\llmeval}{\module{LLMEval}}
\newcommand{\checkereval}{\module{CheckerEval}}

\usepackage{amssymb}
\usepackage{pifont}
\title{OpenFactCheck: Building, Benchmarking Customized Fact-Checking Systems and Evaluating the Factuality of Claims and LLMs}

\author{Yuxia Wang\textsuperscript{1} \quad Minghan Wang\textsuperscript{2} \quad Hasan Iqbal\textsuperscript{1} \\ 
\textbf{Georgi Georgiev}\textsuperscript{3} \quad 
\textbf{Jiahui Geng}\textsuperscript{1} \quad \textbf{Iryna Gurevych}\textsuperscript{1} \quad \textbf{Preslav Nakov}\textsuperscript{1}\\
\textsuperscript{1}MBZUAI \quad \textsuperscript{2}Monash University \quad  \textsuperscript{3}Sofia University \\
  \texttt{\{yuxia.wang, hasan.iqbal, preslav.nakov\}@mbzuai.ac.ae}
}

\begin{document}
\maketitle
\begin{abstract}
The increased use of large language models (LLMs) across a variety of real-world applications calls for mechanisms to verify the factual accuracy of their outputs.
Difficulties lie in assessing the factuality of free-form responses in open domains.
Also, different papers use disparate evaluation benchmarks and measurements, which renders them hard to compare and hampers future progress.
To mitigate these issues, we propose \textbf{\ofc}, a unified framework for building customized automatic fact-checking systems, benchmarking their accuracy, evaluating factuality of LLMs, and verifying claims in a document.
\ofc consists of three modules:
(i) \custchecker allows users to easily customize an automatic fact-checker and verify the factual correctness of documents and claims,
(ii) \llmeval, a unified evaluation framework assesses LLM's factuality ability from various perspectives fairly, and (iii) \checkereval is an extensible solution for gauging the reliability of automatic fact-checkers' verification results using human-annotated datasets.
Data and code are publicly available at \url{https://github.com/yuxiaw/openfactcheck}.

\end{abstract}

\section{Introduction}




Large language models (LLMs) have demonstrated impressive capabilities in generating naturally-sounding answers over a broad range of human inquiries.
However, \gptfour~\citep{gpt4} and other text generation models still produce content that deviates from real-world facts~\citep{bang2023multichatgpt, wang2024factuality-survey}. 
This degrades the system performance and undermines its reliability, representing a significant bottleneck in the deployment~\citep{chuang2023dola, geng2023survey}.
We wonder about the percentage of factually incorrect content generated by LLMs and seek to identify the types of topics and scenarios in which LLMs are most prone to producing false claims.
 
Many studies have explored evaluating and improving the factuality of LLMs~\cite{lee2022factuality, chuang2023dola, shi2023trusting, chen2023felm}.
Two challenges have been identified for evaluation: (1) it is difficult to assess the factuality of open-domain free-form responses, and (2) different papers use different evaluation datasets and measurements, rendering them hard to compare and hampering future progress~\cite{wang2024factuality-survey}.
To mitigate these issues, we introduce a fact-checking framework \textbf{OpenFactCheck}.

\begin{figure*}[t!]
	\centering
        \includegraphics[scale=0.36]{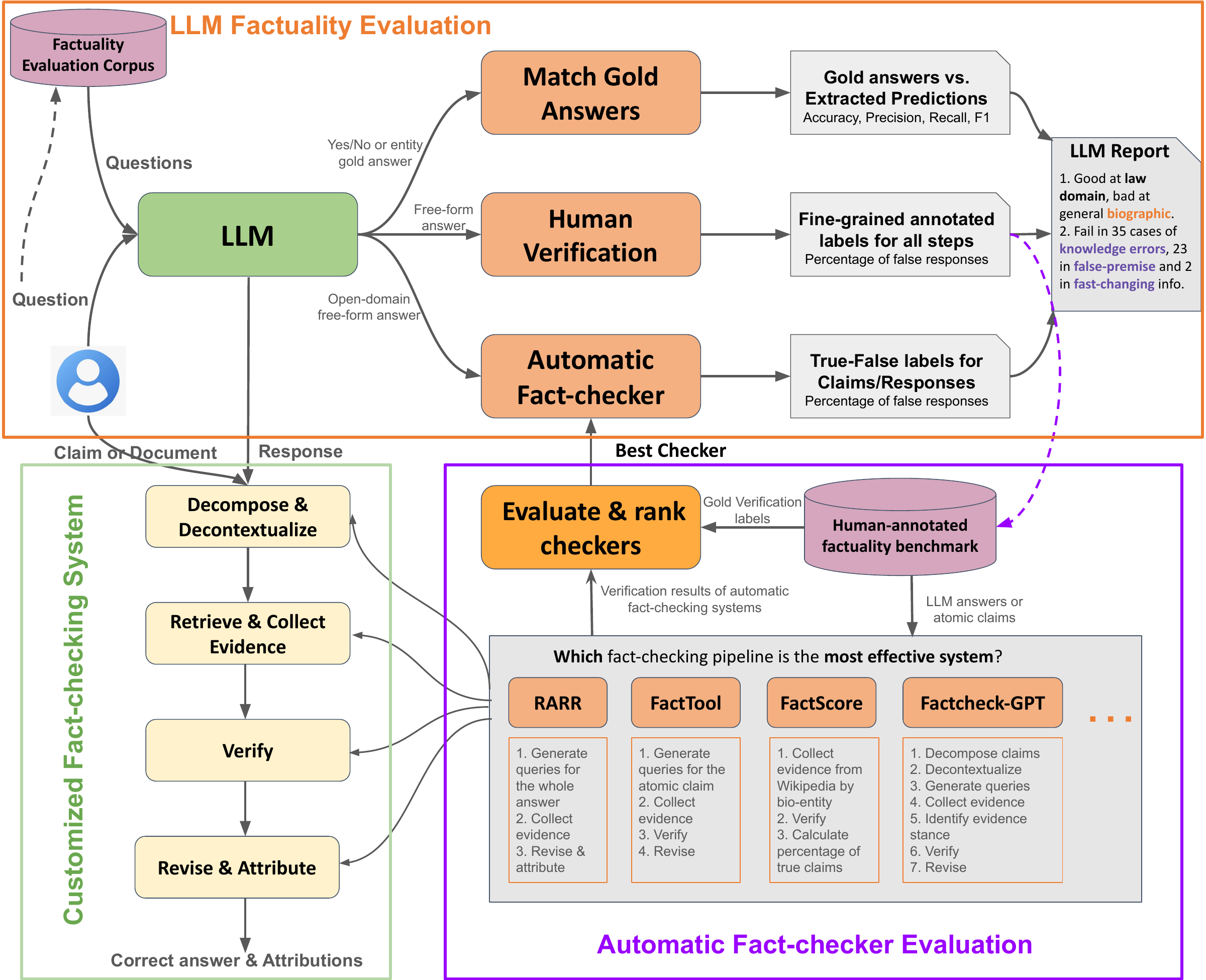}
	\caption{Overview of the \ofc framework with three modules. Green \custchecker: a customized fact-checker to identify factual errors given the outputs of LLMs. Orange \llmeval: a unified LLM factuality evaluator to assess the LLM factual ability from different aspects, and then to produce a report highlighting the domains and types of factual errors the model frequently makes, along with the improvement advice. Purple \checkereval: a fact-checker evaluator and leaderboard to encourage the development of advanced checkers in terms of performance, time and costs.}
	\label{fig:framework}
\end{figure*}

It includes three modules as shown in \figref{fig:framework}. \custchecker allows users to customize an automatic fact-checking system and to verify free-form documents to alleviate the first problem. 
\llmeval is a unified LLM factuality evaluation module. It applies seven factuality-specific benchmarks to assess the LLM factuality ability from different aspects and then produces a report to illustrate the weakness and offer improvement advice. This tackles the second challenge.
We further incorporate \checkereval that assesses the verification accuracy of automatic fact-checking systems, equipped with a leaderboard in terms of accuracy, latency, and costs, aiming to encourage the development of advanced automatic fact-checking systems. 

The three modules are designed to mutually reinforce one another. 
The human-annotated verification results derived from \llmeval can be used as the benchmark for evaluating the accuracy of automated fact-checkers. Simultaneously, the most effective checker identified in \checkereval can be deployed for automated fact-checking tasks.
Each fact-checker in \checkereval can be an implementation in \custchecker. Complex user inquiries may be considered as potential candidates included into the factuality assessment dataset utilized in \llmeval.

Users can tailor their checkers according to their specific needs, such as domain specialization, cost-effectiveness, or rapid processing, and identify factual errors for both human-written text (a claim or document) and the outputs of LLMs.
LLM researchers and practitioners can directly input their LLM responses to the \llmeval, where responses are answers based on our question set.
Subsequently, we conduct evaluations to assess the model's factuality and to generate a report analyzing the model performance from multiple aspects, including snowballing hallucination, self-awareness of unknown knowledge, answering open-domain and latest-information questions.
Similarly, developers who seek to evaluate and compare the efficacy of their fact-checking systems to other ones fairly can upload their checker's verification outcomes to \checkereval. 
Then, our system will show the ranking information in the leaderboard after evaluating under the same measurements.

To sum up, this work investigates three questions:
\begin{compactitem}
\item how to effectively identify factual errors in a document (e.g., an LLM response);
\item how to systematically evaluate the factuality ability of an LLM;
\item which automatic fact-checker is the best, and which component dominates the final verification accuracy.
\end{compactitem}
We initiate an open-source project and develop a preliminary implementation of the three modules, which is anticipated to serve as a stepping stone to facilitate future endeavors in this domain.
We encourage extensive implementation of unique, effective, and robust claim processors, retrievers and verifiers within fact-checking pipelines, collections of challenging questions where LLMs are prone to making factual errors, and human-annotated verification examples. We believe that this will help to promote future research on LLM factuality.  

\section{Background}
In this section, we first analyze existing automatic fact-checking systems, followed by a review of previous approaches to evaluating the factuality of LLMs and the accuracy of fact-checking systems.

\subsection{Fact-checking Systems}
Fact-checking is the task of assessing whether claims made in writing are false or true.
Many recent papers have described automatic fact-checking systems used to evaluate the factuality of LLM responses, such as \rarr~~\cite{gao2022attributed}, \factscore~\cite{min2023factscore}, \factool~\cite{chern2023factool}, \cove~\cite{dhuliawala2023chain}, \factcheckgpt~\cite{wang2023factcheck} and \fire~\cite{xie2024fire}.
Each checker has unique characteristics designed for specific domains or scenarios.
\rarr verifies a document as a whole and can generate an attribution report to explain factual errors. \factscore retrieves evidence from an offline Wikipedia dump mainly for the biography. \factool is friendly to users with low latency, \cove completely depends on the capability of LLMs. \factcheckgpt has a fine-grained pipeline to localize intermediate errors, and \fire is time- and cost-efficient. 
Unlike the above work, our aim is to enable the easy customization of a fact-checker according to users' requirements and application scenarios, e.g., offline settings with a limited budget, by simply clicking dropdowns to choose offline retrievers and verifiers supported by small models without calling APIs.

Despite different designs and implementations of various checkers, they generally consist of three modules: (1) \emph{claim processor}, which extracts context-independent atomic claims from a document, (2) \emph{evidence retriever}, which searches related passages from the Internet or database, and then ranks them by relevance, and (3) \emph{verifier}, which determines the claim/document factuality based on the collected evidence~\citep{guo-etal-2022-survey, li2023selfchecker, wang2024factuality-survey}.
To this end, we first unify different fact-checking systems into a unified pipeline with the three modules. Then, we can select a developed module from various implementations or developers can implement one by themselves.
In addition, the framework supports easy migration of existing fact-checking systems to our pipeline.
However, the verification results of automatic fact-checkers are not necessarily accurate.
How to evaluate and improve the accuracy of automated fact-checkers is critical, since the accuracy serves as a confidence and reliability signal for the verification results.

\subsection{Evaluation of Fact-Checking Systems} 
How accurate are current fact-checking systems? Can they effectively serve as proxies for evaluating the factual accuracy of language models?
Existing automatic fact-checking studies often first collect a set of human-annotated (LLM response, extracted claims, factuality of the claims) in \figref{fig:factoolqa-example}, and then quantify the effectiveness of their systems by comparing the final verification results (i.e., whether a claim or a document is factually \textit{true or false}) to human-annotated labels~\cite{min2023factscore, chern2023factool, dhuliawala2023chain}.
Recent work on long-form factuality in LLMs also demonstrates a statistically significant correlation between the outputs by automatic fact-checkers and labels by human annotators~\citep{wei2024longformfactuality}.
Thus, we merge four human-annotated LLM factuality datasets including \factoolqa, \felmwk, \factcheckbench, and \halueval, and then compare them to the results of automatic fact-checkers to assess the performance of fact-checking systems.

\subsection{LLM Factuality Evaluation}
There are subtle differences between evaluating LLM's general performance and factuality.
Question answering (QA) datasets over various domains are always used for general performance evaluation~\citep{dan2021mmlu}. The focus is on judging whether the question is answered correctly. If the model's response contains the correct answer, it counts; otherwise, it is void even though the response presents all facts.
While research on factuality primarily focuses on whether the response aligns with world knowledge, even if some statements were irrelevant to the question.

Therefore, instead of using datasets for general performance assessment, we selected seven datasets that are specifically collected for factuality evaluation.
They were selected to cover as diverse potential factual errors as possible, including vulnerabilities of snowballing hallucinations, awareness of self-uncertainty, robustness to false-premise questions, fresh questions with answers changing fast over time, and free-form responses spanning distinct domains, topics, and tasks.


\paragraph{Summary} 
Observations above motivate us to integrate these three components into one framework to facilitate (i) users to flexibly configurate an automatic fact-checking system to verify the factuality of claims and documents, (ii) LLM developers to evaluate LLM's factuality under the same measurement scale, and (iii) researchers to assess the fact-checkers reliability under fine-grained annotated benchmarks.


\section{\ofc}
\ofc's design emphasizes two principles: (i) customizability and extensibility for both users and developers, and (ii) compatibility with existing methods and datasets.
It consists of three modules: \custchecker, \llmeval, and \checkereval. Below, we present the detailed design and implementation of each component.

\subsection{\custchecker}
\label{sec:custchecker}

\custchecker allows users to customize a fact-checking system by selecting a claim processor, a retriever, and a verifier. Current version supports the following fact-checking systems: \rarr, \factool and \factcheckgpt~\citep{gao2022attributed, chern2023factool, wang2023factcheck}.
Users can input either human-written text or outputs of LLMs,
and then the fact-checker defined above will process and detect factual errors, showing the verification results including evidence, judgment, and explanations (see \figref{fig:page1}). 

\begin{table*}[t!]
  \centering
  \resizebox{\textwidth}{!}{
  \begin{tabular}{l|p{10cm}|p{6.5cm}|l r}
    \toprule
    \textbf{Dataset}$\downarrow$ & \textbf{The Ability to Evaluate} & \textbf{Domain} & \textbf{Error} & \textbf{Size} \\
    \midrule
    \textbf{\snowball} & Snowballing hallucination when model immediately output & Math, history, graph search & Type 2 & 1,500 \\
    \textbf{\selfaware} & Understand their own limitations on the unknowns & Biology, philosophy, psychology, history & Type 1,3 & 3,369 \\
    \textbf{\freshqa} & Answer questions changing fast over time or with false premises & Sports, entertainment, history, technology & Type 3 & 600 \\ 
    \textbf{\factoolqa} & Respond knowledge-based questions & History, geography, biology, science & Type 1 & 50 \\
    \textbf{\felmwk} & Answer world-knowledge questions & History, biology, geography, sports & Type 1 & 184 \\
    \textbf{\factcheckbench} & Answer open-domain, false-premise questions & Technology, history, science, sports & Type 1,2 & 94 \\
    \textbf{\factscorebio} & Generate detailed biographies & Biography & Type 1,3 & 683 \\
   \midrule
    \textbf{Total} & LLM factuality against world knowledge & 482 domains, top20 accounts for 70\% & Type 1,2,3 & 6,480 \\
    \bottomrule
  \end{tabular}
  }
  \caption{\textbf{\factqa}: factual vulnerability, domain, potential error type and size across seven component datasets.}
  \label{tab:dataset-statistics}
\end{table*}

\paragraph{Configurable Pipeline} 
We unify diverse fact-checking systems as a procedure with three steps, abstracted into three classes: a claim\_processor, a retriever, and a verifier. 
The instances of the three classes are sequentially linked into a pipeline, solving the following tasks: (i) decomposing a document into atomic claims, (ii) collecting relevant evidence passages given a claim, and (iii) making a true/false judgment given both the claim and the evidence as input. 
We refer to them as task solvers.
The implementation of a task solver can be flexible, just ensuring that the input and the output are aligned with the abstract class definitions. For example, evidence can be retrieved by calling SerperAPI or by searching Wikipedia using BM25, but we must return a list of relevant passages given an input claim.

Moreover, task solvers in our pipeline are not hardcoded, but can be configured through a \textit{yaml} configuration file. Thus, users can combine task-solver implementations from different systems (e.g., using \factcheckgpt's claim processor, \rarr's retriever, and \factool's verifier) and start the verification from any step. For example, users can start from the step of retrieval when the input does not need decomposition.


This functionality is achieved by a message-passing mechanism, where a \textit{success flag} is used to indicate whether the current task solver successfully executes and returns the expected output.
The success flag passes through the configured solver order of the pipeline, guaranteeing that the output of the preceding solver fits the input for the current solver, otherwise error warning will be issued. 
Practically, the input and the output parameter names for the task solvers are defined in the configuration file. To link different solvers into a pipeline, one only needs to ensure that the current solver output name matches the input name of the succeeding solver. A dictionary format \textit{fact-checking-state} is kept throughout the pipeline to store all information in the verification.

\paragraph{Extendable Architecture} Inspired by Fairseq, our framework is designed to be highly extendable by treating any third-party task solvers as plugins~\citep{ott2019fairseq}.
As long as the developed task solvers adhere to our class interface definitions, they can be imported and used in our framework.

To sum, customizable and extendable nature of \custchecker allows general users to utilize it as an application with web-based user interfaces.
Advanced developers have the flexibility to use it as a library, developing and integrating their solvers. 

\subsection{\llmeval}
\label{sec:llmeval}
We observed that studies assessing language models' factuality or evaluating whether the methods are effective to mitigate model hallucinations use different datasets and metrics.
This makes it difficult to compare, in the same conditions, the factuality of different models as well as to compare the effectiveness of different factuality enhancement approaches.
Moreover, a lot of prior work applied datasets such as MMLU~\citep{dan2021mmlu}, StrategyQA~\citep{geva2021strategyqa} and HotpotQA~\citep{yang2018hotpotqa} to evaluate model's factuality.
These datasets tend to focus on assessing the general performance, rather than factuality.
To this end, we first collect a dataset \textit{\factqa} by gathering a large number of factual questions that probe diverse factual errors and span across a spectrum of domains, to fairly evaluate LLMs' factuality under the same criteria.

\paragraph{Factual Question Collection}
We collect a set of questions by gathering questions from seven existing corpora that is collected deliberately to assess LLM's factuality, including \snowball~\citep{zhang2023snowball}, \selfaware~\citep{yin-etal-2023-large}, \freshqa~\citep{vu2023freshllms}, \factoolqa~\citep{chern2023factool}, \felmwk~\citep{chen2023felm}, \factcheckbench~\citep{wang2023factcheck} and \factscorebio, a total of 6,480 examples shown in \tabref{tab:dataset-statistics}, referring to \factqa (see dataset details in \appref{app:factqadatasets}).

To concretely analyze models' vulnerability, we identify three labels for each question from the perspective of the knowledge domain, the topic, and the potential error type if a LLM generates a factually incorrect response. 
So each example includes the following fields: \textit{question}, \textit{domain}, \textit{topic}, \textit{ability to test}, \textit{task} and \textit{source}. Domain and topic are identified using \gptfour based on the (question, reference response).\footnote{We used \gptfour response as a reference response for a question as it is more likely to provide a relevant and correct answer, assisting the identification of domains and topics.} 
Domains involve general, legal, biomedical, clinical, scientific and so on. Given a domain, we further fine-grained topics. 
Three common error types are presented.

\textbf{\ftype: \textit{Knowledge error}} is the most common error, occurring when the model produces hallucinated or inaccurate information due to lacking relevant knowledge or internalizing false knowledge in the pre-training stage or in the problematic alignment process.
However, LLMs do not know what they do not know, sometimes overestimate their capacities and confidently output unknown information, leading to false responses.
Mitigating such errors require: (a) learning and correcting parametric knowledge through the curation of corpora used in pre-training, supervised fine-tuning (SFT) and alignment, (b) augmenting by external knowledge in inference, (c) calibrating models to be aware of unknowns, and (d) configuring the decoding strategies (sample/beam-search, temperature), balancing diversity and accuracy~\cite{zhang2023siren}.

\textbf{\stype: \textit{Over-commitment error}} occurs when the model fails to recognize the falsehoods (or jokes) inherent in the prompt or previously-generated context, and provides an inaccurate or inappropriate response.
The left-to-right generation strategy used by LLMs poses potential risks that LLMs sometimes over-commit to the false premise in the context, even when they recognize they are incorrect~\citep{zhang2023siren}.
To address this issue, engineering better prompts is helpful, such as explicitly instructing models to first detect false premises in the prompt~\citep{vu2023freshllms} and asking the same question in a different way (\textit{Is 10733 a prime number?} $\rightarrow$ \textit{What are the factors of 10733? Let's think step-by-step.}) 

\textbf{\ttype: \textit{Disability error}} happens when the model is unable to search up-to-date information to correctly answer questions whose answers change over time, e.g., \textit{What is today's gas price in New York} (fast-changing).
Retrieving external knowledge and augmenting it in the context would help. 

Note that we do not consider \textit{reasoning errors} that arise when a claim employs flawed reasoning or faulty logic, and \textit{irrelevant error} concerning that the content is unrelated to the question~\citep{chen2023felm}. The former highlights LLM's reasoning ability, which is more reflected in math and reasoning tasks, and the latter has more to do with response's helpfulness or human preference. 
They are important in LLM evaluation, and may implicitly influence factuality, but we will first focus on explicit causes, leaving the implicit for future work. 

\begin{table}[t!]
\centering
\resizebox{0.94\columnwidth}{!}{
\begin{tabular}{l|rrr|r}
\toprule
Dataset $\downarrow$  & \#True & \#False & \#Unknown & Total \\
\midrule
\factoolqa & 177 & 56 & 0 & 233 \\
\felmwk & 385 & 147 & 0 & 532  \\
\factcheckbench & 472 & 159 & 47 & 678 \\
\midrule
\halueval & 3,692 & 815 & 0 & 4,507 \\
\bottomrule
\end{tabular}
}
\caption{The number of true, false claims and unknown (no-enough-evidence or opinions) for \factoolqa, \felmwk and \factcheckbench, the number of responses for \halueval (no claim-level labels).}
\label{tab:factbench-statistics}
\end{table}

\paragraph{Evaluation Measurement}
For questions that can be answered by Yes/No or have a short gold answer, we perform exact matching between the model responses and the gold standard answer to judge whether the response is factually correct or not, and then to calculate accuracy, such as for \snowball and \selfaware.
For \freshqa, we use the \textit{FreshEval} proposed in \citet{vu2023freshllms} to evaluate the correctness of model's responses, in which few-shot in-context learning based on \gptfour is applied. We use the strict evaluation criterion which considers an answer to be correct only if all the claims in the response are factually true and also up-to-date.

For open-domain questions from the other four datasets with free-form and long responses, there are no gold standard answers. We use automatic fact-checking systems augmented with retrieved world-knowledge evidence to judge the correctness at the claim-level as well as the document level.

\subsection{\checkereval}
\label{sec:checkereval}
Automatic fact-checking systems aim to identify whether a claim or a document is factually correct or not with/without references, but the results are not necessarily correct.
To assess the accuracy of automatic fact-checkers, we gather four LLM factuality benchmarks with human-annotated factual labels for three levels of granularity text: claims/segments/documents given (question, LLM response) pairs, including \factoolqa, \felmwk, \factcheckbench and \halueval as shown in \tabref{tab:factbench-statistics}, see examples in Figures~\ref{fig:factoolqa-example} and \ref{fig:halueval-example}. 
We refer to them as \factbench. We highlight that \factqa is a question set without answers and human-annotated claims, while \factbench is a benchmark with annotations. 
We use precision, recall, and F1-score with respect to the \textit{True} or \textit{False} claim/document to evaluate the effectiveness of fact-checking systems.

This method regards the system as a whole, only assessing the final verification results, i.e., whether a claim or a document is true or false.
The evaluation of intermediate results throughout the fact-checking pipelines will be incorporated in future updates, to localize which step eventually results in the erroneous factual judgment for claims. 
\section{Experiments}
\label{sec:exp}
We first evaluate the factuality of three LLMs, and then we assess the accuracy of different automatic fact-checking systems in multiple settings. 


\begin{table}[t!]
\centering
\resizebox{\columnwidth}{!}{
\begin{tabular}{lccr}
\toprule
Dataset $\downarrow$  & \llamatwo 7B & \llamatwo 13B & \gptfour \\
\midrule
\factoolqa & 127.3 & 129.5 & 39.5 \\
\felmwk & 131.0 & 125.5 & 62.8 \\
\factcheckbench & 152.7 & 143.9 & 117.3 \\
\midrule
\factqa & 132.1 & 121.9 & 82.2 \\
\bottomrule
\end{tabular}
}
\caption{Word-level length for responses of \llamatwo 7B, 13B and \gptfour over the datasets: \factoolqa, \felmwk, \factcheckbench and \factqa.}
\label{tab:response-length}
\end{table}

\begin{table*}[t!]
\centering
\resizebox{\textwidth}{!}{
\begin{tabular}{c | cccc | rrrr | cc}
\toprule
Dataset $\rightarrow$ & \multicolumn{4}{c|}{\textbf{\snowball}} & \multicolumn{4}{c|}{\textbf{\selfaware}} & \multicolumn{2}{c}{\textbf{\freshqa}} \\
Model $\downarrow$  & Primality  & Senator &  GraphConnection  & Full-set & Precision & Recall & Accuracy  &  F1-score & Accuracy & Perc\_valid\\
\midrule
\llamatwo 7B  & 5.6\%  & 20.4\% & 17.4\% & 14.5\% & 69.7\% & 30.3\% & 74.6\% & 42.0\%  & 28.3\% & 93.2\% \\  
\llamatwo 13B & 0.0\%  & 9.4\%  & 32.4\% & 19.5\% & 64.9\% & 30.1\% & 73.6\% & 41.2\%  & 29.7\% & 95.5\% \\
\gptfour      & 0.2\%  & 49.0\% & 71.0\% & 34.5\% & 71.7\% & 21.6\% & 73.4\% & 33.2\%  & 39.5\% & 98.3\% \\
\bottomrule
\end{tabular}
}
\caption{LLM factuality evaluation accuracy for \snowball: for each of its three topics as well as on average. Shown are \selfaware precision, recall, and F1-score when the positive label=\textit{unanswerable}, and \freshqa accuracy, as well as percentage of valid assessments.} 
\label{tab:em-results}
\end{table*}

\begin{figure*}[t!]
	\centering
	\includegraphics[scale=0.32]{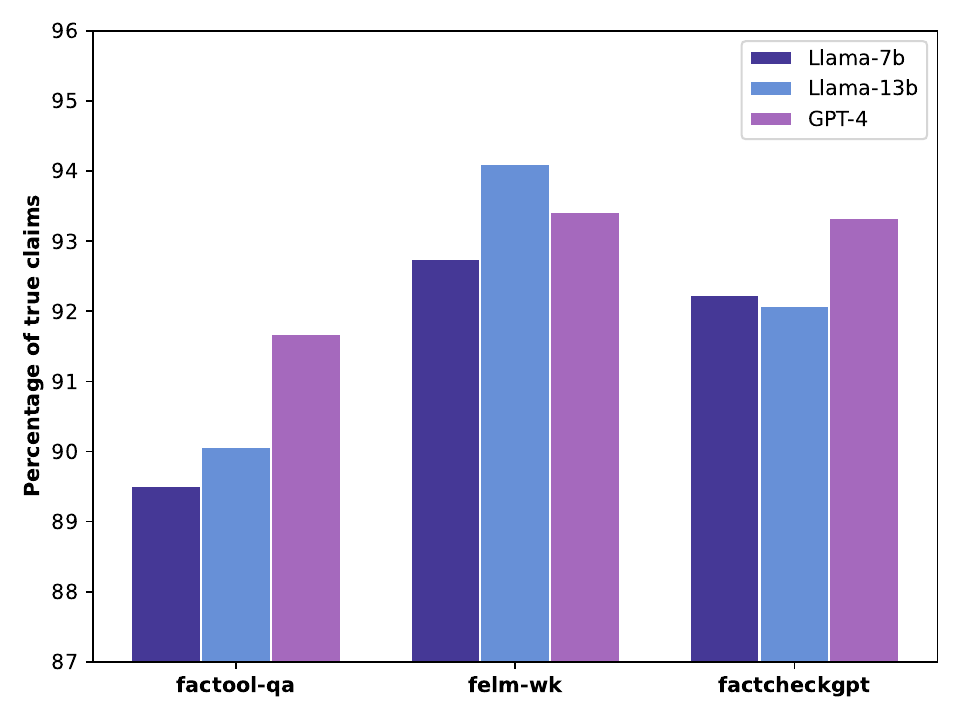}
        \includegraphics[scale=0.32]{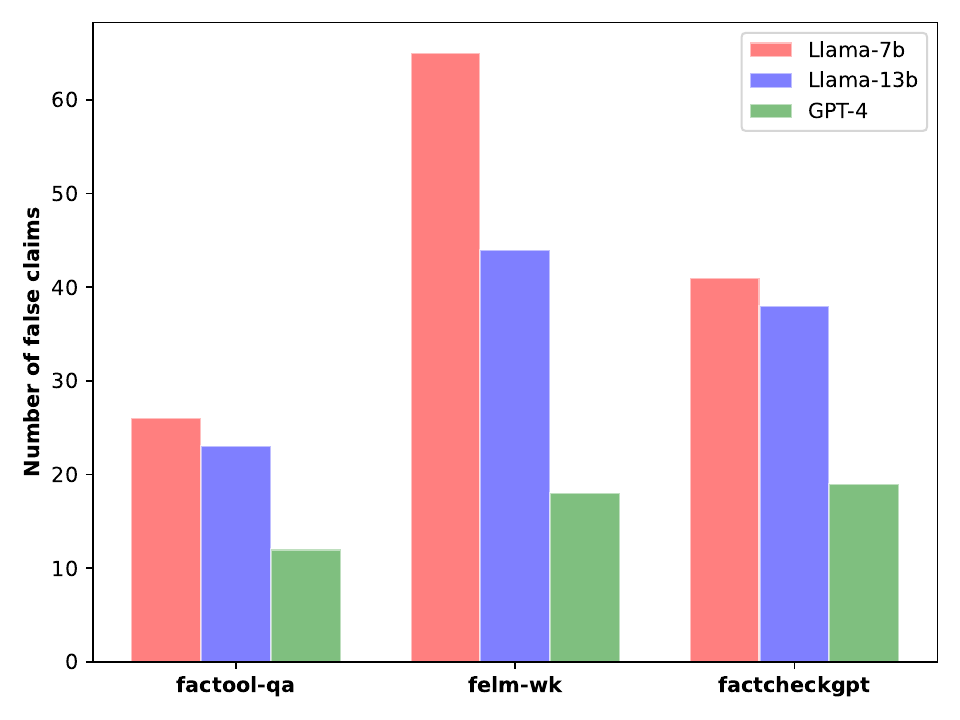}
        \includegraphics[scale=0.32]{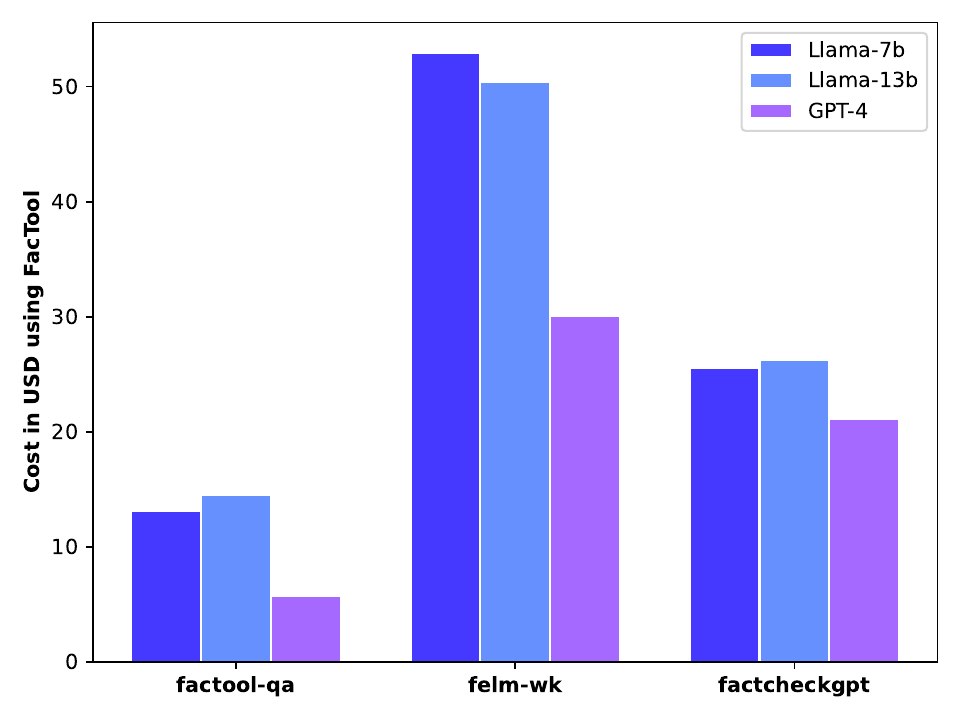}
	\caption{Automatic evaluation results for \llamatwo 7B, 13B and \gptfour responses on datasets of \factoolqa, \felmwk, and \factcheckbench using \factool. \textit{left:} the percentage of true claims, \textit{center:} the number of false claims, and \textit{right:} the cost of using \factool in USD.}
	\label{fig:freeform-evaluation}
\end{figure*}

\subsection{\llamatwo and \gptfour Evaluation}
Based on questions/instructions in \factqa, we collected responses from \llamatwo (7B, 13B) and \gptfour.
Default decoding parameter of \texttt{openai.chat.completions.create()} was used for \gptfour. Decoding parameters of \llamatwo 7B and 13B were taken from the default generation configurations provided in the huggingface transformers library, where we set the \texttt{max\_token} as 512 to avoid the out of memory issue.
As shown in \tabref{tab:response-length}, the responses of \gptfour tend to be shorter than that of \llamatwo.
 
\paragraph{Results and Analysis}
On the \snowball dataset, we observe high error rates: >80\% for \llamatwo and 65.5\% for \gptfour, similar to the results on GPT-3.5-Turbo presented by \citet{zhang2023snowball}. 
However, when justifying previously generated content, \gptfour can identify 87\% of its own mistakes. 
Therefore, in these cases, errors are mostly attributed to the over-committing to the previously generated false context, rather than to large knowledge gaps in LLMs. 
An LLM over-commits to early mistakes, leading to more mistakes that it otherwise would not have made.
Its prevalence in generative models leads to factual errors for simple facts.

\selfaware aims to evaluate LLMs' ability to understand their own limitations and unknowns, identifying unanswerable or unknowable questions.
Higher precision than recall is achieved across three models with regard to unanswerable questions.
This reveals that many truly unanswerable questions are incorrectly recognized as answerable, implying that models are always not aware of what they do not know.
Poor performance on questions with rapidly changing answers (\freshqa) illustrates the inherent challenge of retrieving up-to-date information for LLMs.

We used \factool equipped with Serper and GPT-3.5-Turbo to automatically evaluate the factuality of free-form responses over prompts in \factoolqa, \felmwk, and \factcheckbench.
The results are shown in \figref{fig:freeform-evaluation}, where we can make several interesting observations: 
\begin{itemize}
\item The percentage of true claims is in the range of 89\%-94\%, revealing that the vast majority of claims are verified as true.
\item The questions in \factoolqa are relatively more challenging for the three LLMs to answer correctly than for the other two datasets, leading to a relatively lower percentage of true claims. The apparent lower number of false claims in \factoolqa stems from its smaller dataset size, where 50 is less than 94 and 184.
\item \gptfour has the best factuality performance with a smaller number of false claims and higher percentage of true claims, followed by \llamatwo 13B and then 7B; 
\item The cost for automatic evaluation mainly depends on the number of atomic claims and the price of the backend models used in \factool. It spends \$0.02 for an atomic claim on average. 
\end{itemize}

\textbf{Summary} snowballing hallucination, over-commitment to false premise, difficulty in identifying unknown knowledge and answering with up-to-date information are still challenging issues for LLMs. 
For general open-domain questions, on average less than 10\% of the claims are factually incorrect in LLM responses.
This somehow implies that models may poorly understand instructions and their knowledge scope, but they can correctly generate majority of content. 
This is aligned with the recent finding that what an LLM can generate, it may not understand~\citep{west2023paradox}.
Additionally, it is costly to evaluate open-domain answers even if based on automatic fact-checkers, $\sim$ \$30 for 100 responses based on the cheapest \gptthreepointfive.

\begin{table*}[t!]
\scriptsize
    \centering
    \resizebox{\textwidth}{!}{
    \begin{tabular}{l|c|c|ccc|ccc|ccc|ccc|ccc|ccc}
    \toprule
    \multicolumn{1}{c|}{\multirow{3}{*}{\textbf{Framework}}} & \multicolumn{1}{c|}{\multirow{3}{*} {\textbf{Verifier}}} & \multicolumn{1}{c|}{\multirow{1}{*}{\textbf{Source/}}} & \multicolumn{6}{c|}{\textbf{\factcheckbench}} & \multicolumn{6}{c|}{\textbf{\factoolqa}} & \multicolumn{6}{c}{\textbf{\felmwk}} \\
    & & \textbf{Retriever} & \multicolumn{3}{c|}{\textbf{Label = True}} & \multicolumn{3}{c|}{\textbf{Label = False}} & \multicolumn{3}{c|}{\textbf{Label = True}} & \multicolumn{3}{c|}{\textbf{Label = False}} & \multicolumn{3}{c|}{\textbf{Label = True}} & \multicolumn{3}{c}{\textbf{Label = False}} \\
    & & & Prec & Recall & F1 & Prec & Recall & F1 & Prec & Recall & F1 & Prec & Recall & F1 & Prec & Recall & F1 & Prec & Recall & F1 \\
    \midrule
    Random & -- & -- & 0.79 & 0.43 & 0.56 & 0.18 & 0.52 & 0.27 & 0.79 & 0.56 & 0.66 & 0.28 & 0.54 & 0.37 & 0.71 & 0.52 & 0.60 & 0.26 & 0.43 & 0.32\\
    Always True & -- & -- & 0.81 & 1.00 & 0.88 & 0.00 & 0.00 & 0.00 & 0.76 & 1.00 & 0.86 & 0.00 & 0.00 & 0.00 & 0.72 & 1.00 & 0.84 & 0.00 & 0.00 & 0.00 \\
    Always False & -- & -- & 0.00 & 0.00 & 0.00 & 0.19 & 1.00 & 0.33 & 0.00 & 0.00 & 0.00 & 0.24 & 1.00 & 0.39 & 0.00 & 0.00 & 0.00 & 0.28 & 1.00 & 0.43 \\
    \midrule
    \factscore & \llama3-Inst 8B & Wiki/BM25 & 0.87 & \textbf{0.74} & 0.80 & 0.34 & 0.56 & 0.42 & 0.82 & 0.68 & 0.74 & 0.34 & 0.52 & 0.41 & 0.76 & 0.66 & 0.71 & 0.34 & 0.46 & 0.39 \\
    \factool & \llama3-Inst 8B & Web/Serper & 0.88 & 0.80 & \textbf{0.84} & 0.40 & 0.56 & 0.47 & \textbf{0.93} & 0.38 & 0.54 & 0.32 & \textbf{0.91} & 0.47 & 0.79 & 0.31 & 0.44 & 0.30 & \textbf{0.79} & 0.44 \\
    \factscore & GPT-3.5-Turbo & Wiki/BM25 & 0.87 & 0.67 & 0.76 & 0.31 & 0.60 & 0.41 & 0.82 & 0.58 & 0.68 & 0.31 & 0.59 & 0.40 & 0.77 & 0.71 & 0.74 & 0.36 & 0.43 & 0.39\\
    \factool & GPT-3.5-Turbo & Web/Serper & 0.89 & \textbf{0.74} & 0.81 & 0.37 & 0.62 & 0.46 & 0.92 & 0.59 & 0.72 & 0.39 & 0.84 & 0.53 & 0.78 & 0.62 & 0.69 & 0.35 & 0.54 & 0.43 \\
    \factcheckgpt & \gptfour & Web/SerpAPI & 0.90 & 0.71 & 0.79 & \textbf{0.52} & \textbf{0.80} & \textbf{0.63} & 0.88 & \textbf{0.88} & \textbf{0.88} & \textbf{0.63} & 0.63 & \textbf{0.63} & \textbf{0.81} & \textbf{0.87} & \textbf{0.84} & \textbf{0.55} & 0.44 & \textbf{0.49} \\
    \midrule
    \perplexityai & Sonar-online & Web & \textbf{0.93} & 0.73 & 0.83 & 0.40 & 0.76 & 0.53 & 0.82 & \textbf{0.88} & 0.85 & 0.50 & 0.38 & 0.43 & 0.76 & 0.82 & 0.79 & 0.40 & 0.31 & 0.35 \\
    \bottomrule
    \end{tabular}}
    \caption{\textbf{Verification results} for human-annotated claims in \factcheckbench, \factoolqa, and \felmwk, judging whether or not a claim is factually true or false with external knowledge (Wikipedia or Web articles) as evidence. The implementation of \factcheckgpt usedlangchain AutoGPT. \gptfour refers to \textit{gpt-4-turbo-2024-04-09}.}
    \label{tab:verification}
\end{table*}

\begin{table}[t!]
\centering
\resizebox{\columnwidth}{!}{
\begin{tabular}{l|l|l|c}
\toprule
Fact-Checker $\downarrow$  & Web search (\$) & LLM (\$) & Time (hrs) \\
\midrule
\factool & 2.5 (Serper) & 12.2 (GPT-3.5) & 0.49\\
\factcheckbench & 13.3 (SerpAPI) & 26.6 (\gptfour) & 7.67 \\
\bottomrule
\end{tabular}
}
\caption{Time and USD cost for evaluating the 765 claims in \factoolqa and \felmwk.}
\label{tab:checkercost}
\vspace{-1em}
\end{table}


\subsection{Evaluating Fact-Checking Systems}
We investigate automatic fact-checking systems in three aspects: accuracy, latency, and costs.
Based on annotated factual labels for claims from three benchmarks of \factcheckbench, \factoolqa, and \felmwk, we evaluate the verification performance in multiple settings across different fact-checking frameworks, evidence sources, and verifiers. 

Pipeline and core component modules of different fact-checking frameworks are basically similar, including obtaining atomic claims, collecting evidence and verifying correctness; all thus, while the implementations are different. 
For example, in terms of how to extract atomic claims, \rarr does not include this step. 
\factscore first breaks down a document into paragraphs, and then applies NLTK~\citep{bird-loper-2004-nltk} to split paragraphs into sentences, and then prompts LLMs to decompose to atomic claims (GPT3 was used in the original paper). However, this implementation neglects the decontextualization in the paragraph and in the sentence decomposition, making claims non-independent (e.g., \textit{He is a university professor and the CEO of a tech startup company}).
To mitigate, \factool directly extracts claims based on the document, and \factcheckgpt decontextualizes both sentences and claims based on the document.

\paragraph{Experimental Setup} 
To ensure that all fact-checking systems verify the same sets of annotated claims, we skip the step of extracting atomic claims from the documents. All systems get a claim as an input, and they are expected to output whether or not the claim is true.

Recent fact-checking frameworks such as \factscore, \factool, \factcheckgpt and commercial retrieval-augmented generative models such as \perplexityai are evaluated, with evidence retrieved from Wikipedia articles or web pages, as well as with various LLM-based verifiers that judge the factuality of a claim based on their internal knowledge and retrieved evidence as a reference.

\paragraph{Results and Analysis}
In \tabref{tab:verification}, we observe that automatic fact-checking systems struggle to detect false claims. Across the three datasets we experiment with, it is consistently more arduous for systems to differentiate false claims compared to identifying true ones.
This challenge may arise from the tendency of returning invalid evidence for false claims. 

Retrieving evidence from the web using Serper (Google search engine results) is more effective than sourcing related passages from Wikipedia articles using BM25, given that a wider array of effective evidence is accessible on open web pages for open-domain questions.
The verification accuracy of an LLM-based verifier primarily relies on the capabilities of the LLM and the effectiveness of the prompts used. For instance, the overall performance of \gptfour surpasses that of both \llama-3-8B and \gptthreepointfive, and thus the verification results of \factcheckgpt outperform those of \factool, \factscore and \perplexityai, despite all of them utilizing evidence sourced from the web.
While \factcheckgpt exhibits superior effectiveness, it is associated with considerable latency and substantial costs (see \tabref{tab:checkercost}). 

Latency and cost are largely contingent upon the implementation strategy. 
For instance, \factool adopts asynchronous processing and leverages Serper (\$0.001 per search) in conjunction with \gptthreepointfive, rendering it faster and more economical compared to \factcheckgpt. Notably, \factcheckgpt uses SerpAPI (\$0.015 per search) alongside \gptfour, where the cost of the most affordable \gptfour model is 20 times that of \gptthreepointfive (see \figref{tab:gpt-info}). 

In summary, the efficacy of automated fact-checking systems is fundamentally dependent on implementation factors such as choice of search tool, prompts, and backend LLMs. This is primarily driven by engineering considerations.

\section{Conclusion and Future Work}

We proposed \ofc, a unified, easy-to-use and extensible framework. It supports the customization and evaluation of automatic fact-checking systems and LLM factuality evaluation.
Specifically, \ofc allows general users to check whether a claim and a document are factual or not, and also facilitate LLM practitioners and developers to effectively and efficiently evaluate the factuality of their LLMs from various perspectives, and to assess the accuracy of automatic fact-checking systems. 

Our extensive experiments indicate that more than 90\% of the claims generated by LLMs in response to open-domain questions are factually correct.
Nevertheless, models encounter challenges when addressing some straightforward questions such as \textit{Is 7411 a prime number?} 
This difficulty can be attributed to the fact that LLMs demonstrate weaker comprehension abilities relative to their generation capabilities.
Additionally, prevalent fact-checking systems struggle to identify false claims, with the retrieval of pertinent evidence posing a significant bottleneck. The latency and the cost associated with these systems primarily hinge on implementation strategies.
In the future, we will continue to integrate new techniques, features, and evaluation benchmarks to \ofc to facilitate the research progress of LLM fact-checking.

\section*{Acknowledgements}
We thank Zain Muhammad Mujahid and Osama Mohammed Afzal from MBZUAI for the many fruitful discussions, which have helped us improve this work.
\section*{Limitations}
While \ofc presents a comprehensive framework for factuality evaluation of LLMs, several limitations must be acknowledged:

\paragraph{Evaluation Datasets}
The effectiveness of \ofc is dependent on the quality and diversity of the datasets used for evaluation. While we have integrated multiple datasets to cover a broad spectrum of domains and potential factual errors, the evaluation is still limited by the inherent biases and coverage gaps in these datasets. For instance, some specialized domains may not be adequately represented, potentially affecting the robustness of the evaluation for LLMs in those areas.

Moreover, we only evaluated factuality of \gptfour and \llamatwo (7B, 13B) responses to demonstrate the framework's capabilities, while the \ofc system is designed to be extensible. Future work will incorporate the new SOTA LLMs and diverse datasets to broaden the evaluation. The current evaluation provides a proof-of-concept, and we welcome collaborations to expand this further.

\paragraph{Latency and Costs}
The performance of automatic fact-checking systems integrated within \ofc can vary significantly in terms of latency and operational costs. High accuracy often comes at the expense of increased computational resources and processing time, which may not be feasible for all users, particularly those with limited budgets or time constraints.

\paragraph{Reliance on External Knowledge Sources}
The fact-checking modules depend heavily on external knowledge sources, such as Wikipedia and web search engines. The availability and reliability of these sources can affect the accuracy and completeness of the fact-checking process. Furthermore, the dynamic nature of web content means that the information retrieved may not always be up-to-date.

\section*{Ethical Statement}
The development and deployment of \ofc are guided by a commitment to ethical principles, ensuring that the framework is used responsibly and for the benefit of society:

\paragraph{Transparency and Accountability}
We strive to maintain transparency in the design, implementation, and evaluation of \ofc. The source code and datasets are publicly available, enabling scrutiny and fostering trust within the research community. We encourage users to report any issues or biases they encounter, facilitating continuous improvement.

\paragraph{Bias Mitigation}
Recognizing that biases can exist in both datasets and LLMs, we are dedicated to minimizing such biases in \ofc. By integrating diverse evaluation benchmarks and encouraging the development of fair fact-checking approaches, we aim to reduce the impact of biases on factuality evaluation outcomes.

\paragraph{Social Impact}
By enhancing the factual accuracy of LLMs, \ofc aims to contribute positively to society. Accurate information is crucial for informed decision-making and public discourse. We believe that improving the reliability of LLM outputs can help combat misinformation and support the dissemination of truthful information.

\bibliography{ref}
\bibliographystyle{acl_natbib}
\clearpage
\onecolumn
\section*{Appendix}
\appendix

\section{Factuality Datasets}
\label{app:alldatasets}

\subsection{\factqa Component Datasets}
\label{app:factqadatasets}
\snowball dataset~\cite{zhang2023snowball} comprises three question--answering subsets: primality testing, senator search, and graph connectivity, each with 500 yes/no questions.
They aim to investigate snowballing hallucination when a model immediately outputs an incorrect answer (yes or no) as false generated context. Specifically, they prompt the language model to first output a yes/no answer and then to provide explanations. When the immediate answer is wrong, the model tends to continue to snowball the false statements instead of correcting them.
%

\selfaware~\citep{yin-etal-2023-large} aims to evaluate LLMs' ability to understand their own limitations and unknowns. This is achieved by assessing models' ability to identify unanswerable or unknowable questions. They compiled a collection of 1,032 unanswerable questions from online platforms like Quora and HowStuffWorks. In addition, they gathered 2,337 answerable questions from sources such as SQuAD, HotpotQA, and TriviaQA, resulting in a total of 3,369 questions.

\freshqa~\cite{vu2023freshllms} is composed of 600 natural, open-ended questions, segmented into four primary categories based on the answer's stability: \emph{never-changing}, for answers that rarely alter, \emph{slow-changing}, for those that evolve over several years, \emph{fast-changing}, for answers that shift within a year or less, and \emph{false-premise}, encompassing questions with factually incorrect premises that need to be countered.

FacTool~\cite{chern2023factool} detected factual errors in LLM generations across four different tasks: knowledge-based QA, code generation, mathematical reasoning, and scientific literature review. During model evaluation, they reported both response-level and claim-level accuracy when the responses consist of several claims. We used 50 knowledge-based QA: \factoolqa in \factqa.
 
FELM~\cite{chen2023felm} introduced a benchmark for factuality evaluation of LLMs. This benchmark collects responses generated from LLMs and annotated factuality labels in a fine-grained manner. The dataset consists of 5 categories, with examples per category as follows: 194 math, 208 reasoning, 125 science, 184 world knowledge (wk), and 136 writing recordings. We used 184 world-knowledge questions, referring to \felmwk.

\factcheckbench~\citep{wang2023factcheck} Factcheck-GPT gathered a total of 94 highly challenging questions from sources including Twitter posts, internal brainstorming, and Dolly-15k, encompassing 678 claims.

\factscorebio~\cite{min2023factscore} selected 183 entities, and collected responses from three LLMs including \davinci, \chatgpt, and \pplai, and then annotated factual labels (supported, not-supported and irrelevant) for each atomic claim by humans.
Specifically, if the atomic claim was clearly not related to the prompt, and thus should be removed from the bio without a validation step, they assigned \textit{Irrelevant}. If the claim was relevant, they validated it based on the English Wikipedia, and labeled it either as \textit{Supported} or \textit{Not-supported}. Additionally, the annotators also edited the text to make it factually correct. The annotators were also asked to correct factual errors and to remove the sentence if the information it contains is entirely off.
Their data can be downloaded from \url{https://drive.google.com/drive/folders/1bLHGu_imkZVtX6O0mpZ-G0-4ofTLM1ZA}. They proposed automatic checking based on retriever + LLM + masked LLM calculating the perplexity to determine the factual labels of atomic claims and to calculate the error rate or a FactScore for an LLM.
They further collected responses from 12 LLMs based on another 500 entities and evaluated their factuality using automatic estimators. 
Overall, they labeled 183 biographies * 3 models = 549 examples, and further used 500 * 12 = 6,000 unlabeled examples.

\paragraph{Domain and Topic}
There are 482 unique domains and 4,740 unique topics (unique by lexicons without semantic clustering).
The top-20 domains are shown in \tabref{tab:domain-dist}, accounting for 70\% or 4,523 examples.
Except for the \snowball dataset: 500 examples for each of primality testing, US senator search and graph connectivity-flight search, there are fewer than 11 examples per topic, generally 1--3 examples.

\begin{table}[ht!]
  \centering
  \begin{tabular}{lr|lr}
    \toprule
    \textbf{Domain} & \textbf{Size} & \textbf{Domain} & \textbf{Size}\\
    \midrule 
    History          & 771 & Science          & 143 \\
    Biography        & 683 & Physics          & 136\\
    Mathematics      & 612 & Social Sciences   & 111\\
    Transportation   & 519 & Literature       & 100\\
    Biology          & 259 & Geography        & 87\\
    Philosophy       & 229 & Astronomy        & 82\\
    Technology       & 208 & Economics        & 69\\
    Entertainment    & 191 & Music            & 66 \\
    Psychology       & 169 & Religion         & 63\\
    Sports           & 157 & General Knowledge & 53\\
    \midrule 
    \textbf{Total} & & \multicolumn{2}{r}{\textbf{4,523 (69.8\%)}}\\
    \bottomrule
  \end{tabular}
  \caption{\factqa's top-20 domains and the number of examples from each domain.}
  \label{tab:domain-dist}
\end{table}

\subsection{Other Datasets}
TruthfulQA~\cite{lin-etal-2022-truthfulqa} is a benchmark designed to measure whether a language model is truthful when generating answers to questions and is robustness with respect to false beliefs and misconceptions. It comprises 817 questions spanning 38 categories, including health, law, finance, and politics. The dataset includes both text generation and multiple-choice components, with the multiple-choice questions having a variable number of options. The questions are designed to be ``adversarial'' to test for weaknesses in the truthfulness of language models rather than testing models on a useful task.

CoVe used four datasets~\cite{dhuliawala2023chain}.
One is selected from Wikidata --- listings of people with specific professions born in a certain city (56 questions), and the other one is listing works from specific categories based on Wiki-category (55 questions), e.g., \emph{Name some Mexican animated horror films} or \emph{Name some Endemic orchids of Vietnam}.
The third dataset consists of 418 questions sampled from the MultiSpanQA test set with shorter answers per span (up to three tokens per item).
They also used the dataset of generated biographies from FactScore.


SelfCheckGPT~\cite{manakul2023selfcheckgpt} generated synthetic Wikipedia articles about individuals/concepts from the WikiBio dataset~\cite{lebret2016generating} using GPT-3, followed by manual annotation to assess the factuality of each passage at the sentence level. 

HaluEval~\cite{li2023halueval} comprises 5,000 general user queries accompanied by ChatGPT responses and 30,000 specialized examples from three distinct tasks: question answering, knowledge-grounded dialogue, and text summarization.

Self-Contradictory~\cite{niels2023selfcontradictory} constructed a dataset by sampling language model responses about various topics, consisting of 360 text descriptions covering 30 diverse topics.

FACTOR~\cite{muhlgay2023factor} introduced a framework that automatically transforms a factual corpus of interest into a benchmark for evaluating an LM's factuality, and created two benchmarks using this framework: Wiki-FACTOR and News-FACTOR. Wiki-FACTOR is based on the Wikipedia section of The Pile's validation split and consists of 2,994 examples, while News-FACTOR is based on Reuters articles extracted from the RefinedWeb Dataset and consists of 1,036 examples. 

HELM~\cite{liang2023holistic} is a living benchmark designed to enhance the transparency of language models. The dataset implements a multi-metric approach, encompassing various user-facing tasks, domains, and languages. It includes a core set of scenarios and metrics, covering tasks such as question answering, information retrieval, summarization, and toxicity detection across different domains and languages. The dataset is intended to be continuously updated with new scenarios, metrics, and models, and all raw model prompts and completions are released publicly for further analysis.

\subsection{\factbench Examples}
Given an example in \factoolqa, \felmwk and \factcheckbench, it includes a prompt, a response generated by GPT-3.5-turbo and human annotations: a list atomic claims decomposed based on the response, a list of factual labels [True, False, $\cdots$] corresponding to each atomic claim, and the factual label for the response, as shown in \figref{fig:factoolqa-example} below:

\begin{figure}[ht!]
    \centering
    \includegraphics[width=\linewidth]{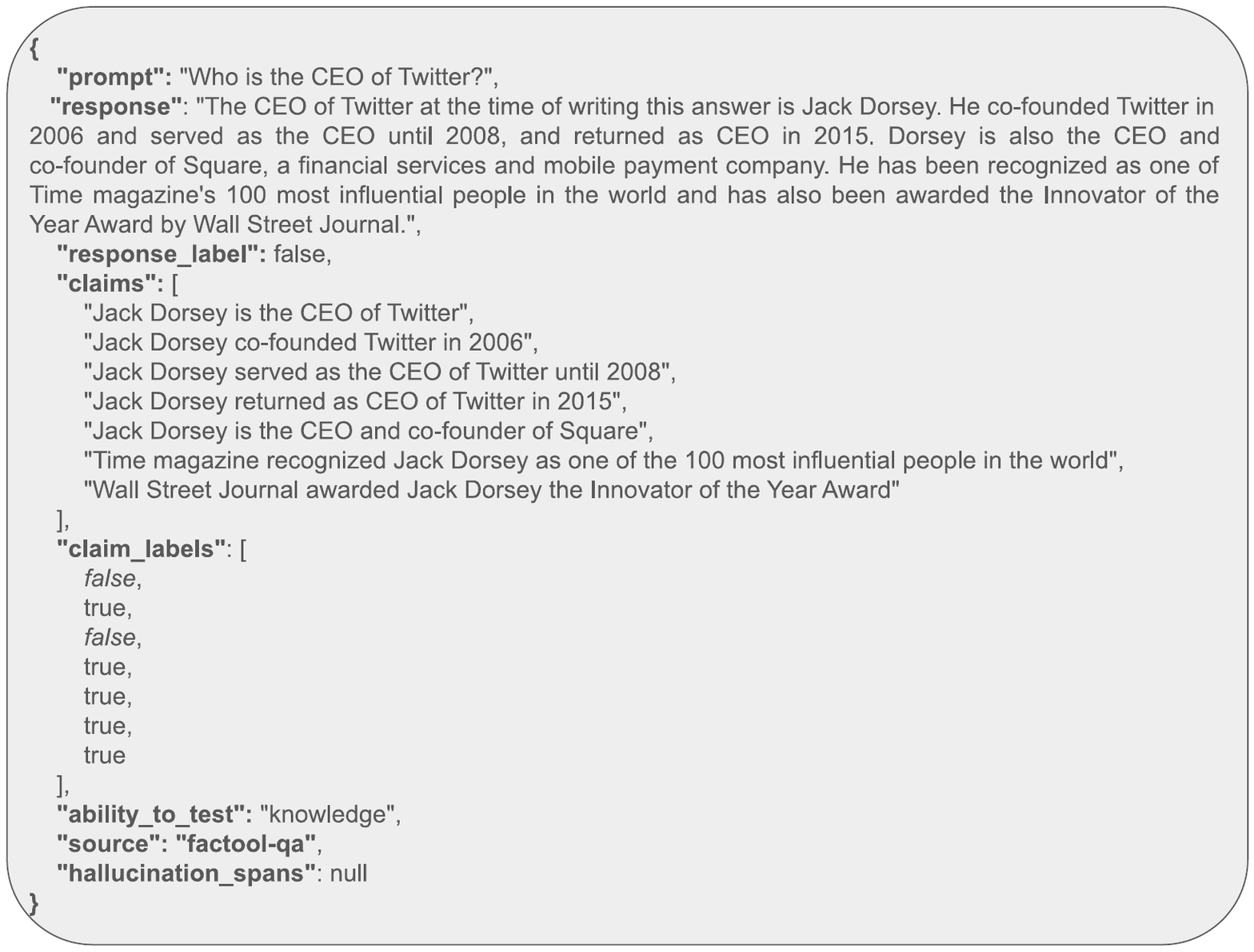}
    \caption{An annotated \factoolqa example sampled from \factbench.}
    \label{fig:factoolqa-example}
\end{figure}

Given an example in \halueval, it includes a prompt, a response by GPT-3.5-turbo and human annotations: a list of false segments/spans extracted from the response, and the factual label for the response as below:
\begin{figure}[ht!]
    \centering
    \includegraphics[width=\linewidth]{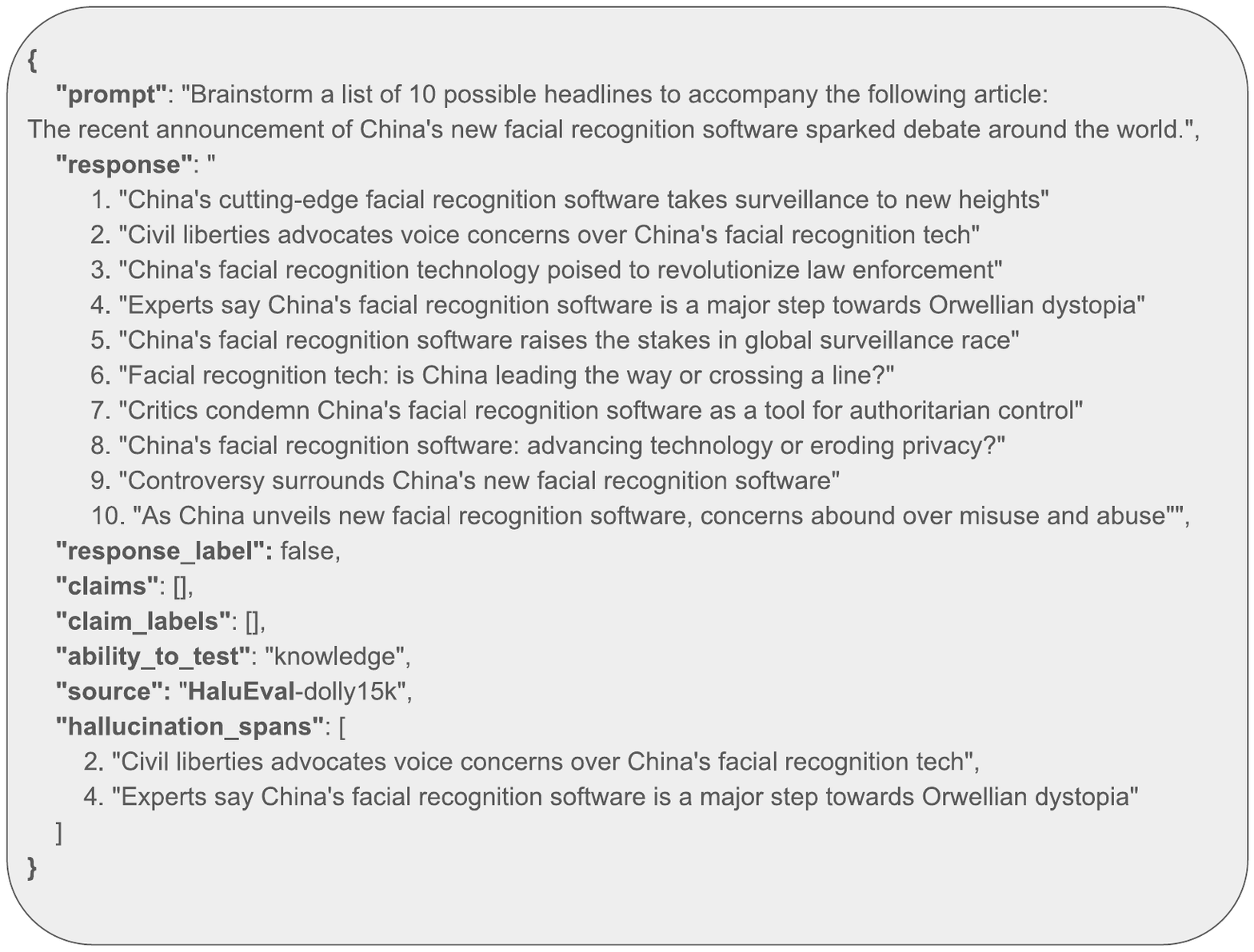}
    \caption{An annotated \halueval example sampled from \factbench.}
    \label{fig:halueval-example}
\end{figure}

\clearpage
\section{Experiments}



\begin{figure*}[ht!]
	\centering
        \includegraphics[scale=0.5]{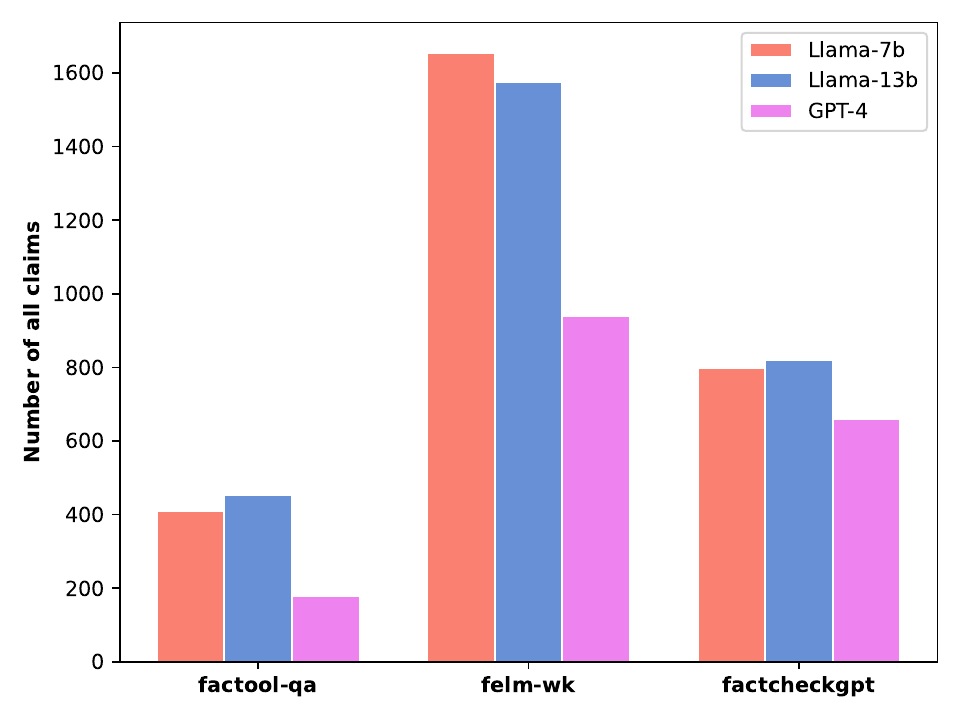}
	\caption{The number of extracted atomic claims using \factool across responses of \llamatwo 7B, 13B, and \gptfour.}
	\label{fig:numclaims}
\end{figure*}

\paragraph{Claim Statistics of Three LLMs' Responses}
\figref{fig:numclaims} shows the number of atomic claims extracted from \llamatwo 7B, 13B and \gptfour responses elicted from the prompts of \factoolqa, \felmwk and \factcheckbench.
There are 50, 184 and 94 prompts in the three datasets respectively, decomposing approximately into 400, 1,600 and 800 atomic claims for \llamatwo responses.
The answers of \gptfour are generally shorter, resulting in 200, 800, and 600 claims.
Note that the three original datasets were annotated with factual labels for GPT-3.5-Turbo responses, and there are 233, 532 and 678 claims for \factoolqa, \felmwk and \factcheckbench.

\begin{figure}[ht!]
    \centering
    \includegraphics[width=\linewidth]{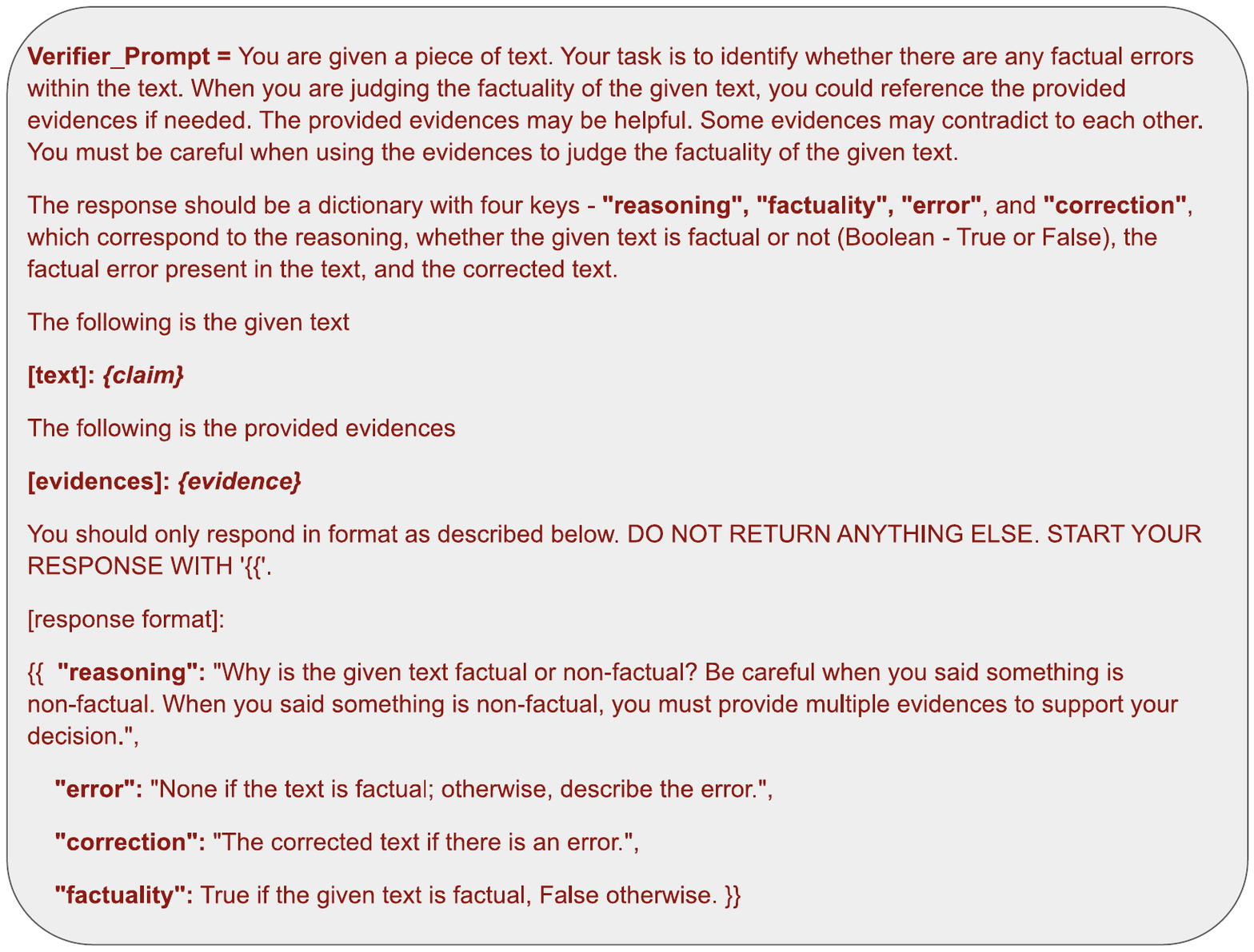}
    \caption{LLM-based verifier prompt example.}
    \label{fig:verifier-prompt}
\end{figure}

\paragraph{LLM-based Verification Prompt} \figref{fig:verifier-prompt} shows a prompt used to verify whether a claim is true or false based on the collected evidence.

\paragraph{OpenAI GPT Models}
\tabref{tab:gpt-info} shows the price, the input context window and the maximum output tokens for two GPT models: \textit{gpt-3.5-turbo-0125} and \textit{gpt-4-turbo-2024-04-09}.
The price of the latter is 20 times the price of the former.
\begin{table*}[ht!]
\centering
\resizebox{\textwidth}{!}{
\begin{tabular}{l|c|c|c|c}
\toprule
Model & Input & Output & Input context window & Maximum output tokens \\
\midrule
\textit{gpt-3.5-turbo-0125} & \$0.5 / 1M tokens & \$1.5 / 1M tokens & 128k & 4096 \\
\textit{gpt-4-turbo-2024-04-09} & \$10 / 1M tokens & \$30 / 1M tokens & 16k & 4096 \\
\bottomrule
\end{tabular}
}
\caption{Price, input context window, and maximum output tokens for two GPT models.}
\label{tab:gpt-info}
\end{table*}

\clearpage
\section{\ofc Demo}
Based on this framework, we initiate a \ofc demonstration. It was implemented with a Python server, a web user interface, and a database, deployed via AWS.
The Python back\-end can also be used as a Python toolkit, allowing easy and flexible development.
This demo was further optimized and presented in \citet{hasan2024openfactcheck}.

\subsection{\custchecker Pseudo Code}
\begin{figure}[ht!]
\centering
\begin{minted}{python}
def claim_processor(document: str) -> List[str]:
    # FactScore
    paragraphs = documents.split("\n")
    sentences = [NLTK(para) for para in paragraphs]
    claims = [call_LLM(sentence, prompt="decompose into atomic claims") for sentence in sentences]

    # FacTool
    claims = call_LLM(document, promot="extract context-independent atomic claims based on the document")
    
    return claims
    
def retriever(claim: str, database: DB, retrieval_strategy: obj, search_api_key: str) -> List[str]:
    # offline DB dump
    evidence = retrieval_strategy(claim, database)
    
    # online web pages by calling API
    evidence = serper_or_serpapi(claim, search_api_key)

    return evidence

def verifier(claim: str, evidence: List[str]) -> bool:
    # call LLMs
    factual_label = call_LLM(claim, evidence, prompt="based on the evidence and your own knowledge, determine whether the claim is true or false.")

    # use NLI models
    stance2factual = {
        "entailment": true,
        "contradiction": false,
        "neutral": "not enough evidence"
    }
    stances = [nli(evid, claim) for evid in evidence]
    majority_stance = majority_vote(factual_labels)
    factual_label = stance2factual[majority_stance]
    
    return factual_label
\end{minted}
    \caption{Pseudo code for the three modules in \custchecker.}
    \label{fig:python_code}
\end{figure}


\subsection{Web Client}
\label{sec:webclient}
We develop a web client based on Streamlit, consisting of four interfaces, with each corresponding to one of the three modules, along with a leaderboard, to enhance the user interaction. 
The design principle is to invoke these functional modules in the form of third-party applications, avoiding excessive intervention in the system's architecture. This makes \ofc a three-in-one to users as a library, a command-line toolkit, and a web application.

\begin{figure}[ht!]
    \centering
    \includegraphics[scale=0.33]{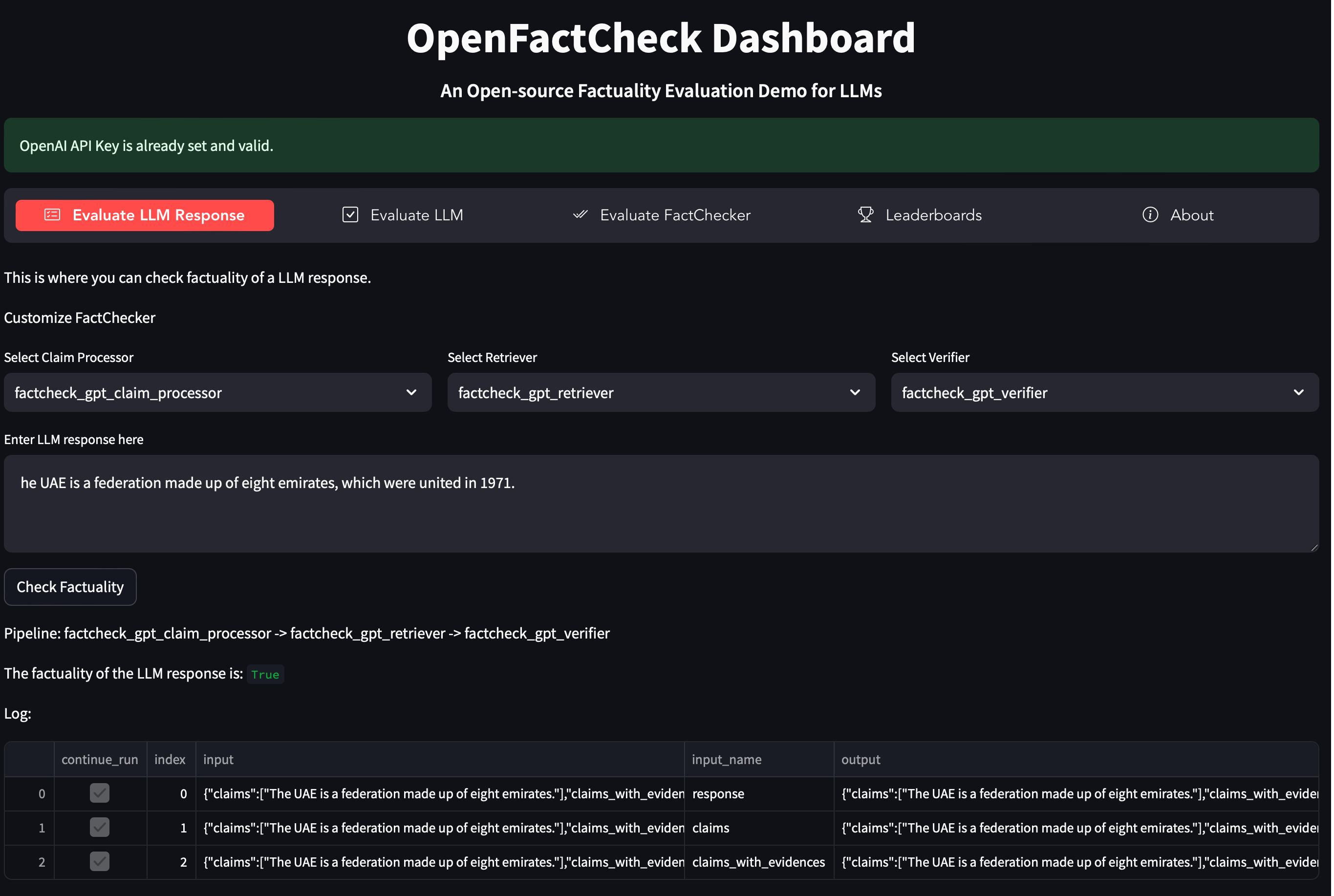}
    \caption{The interface of the Customized Fact-checking System page. The response ``\emph{The UAE is a federation made up of eight emirates, which were united in 1971}'' is a random example for demonstration purposes. We can see the final \emph{False} judgment and the intermediate results.}
    \label{fig:page1}
\end{figure}

As \textit{\custchecker} interface shown in \figref{fig:page1},
users can freely select different combinations of claim processors, retrievers, and verifiers. When given an input document or claim, the \custchecker backend (\S\ref{sec:custchecker}) executes the fact-checking pipeline. The final verification results and the intermediate processing outcomes are presented on the page for reference. 

\begin{figure}[ht!]
    \centering
    \includegraphics[scale=0.33]{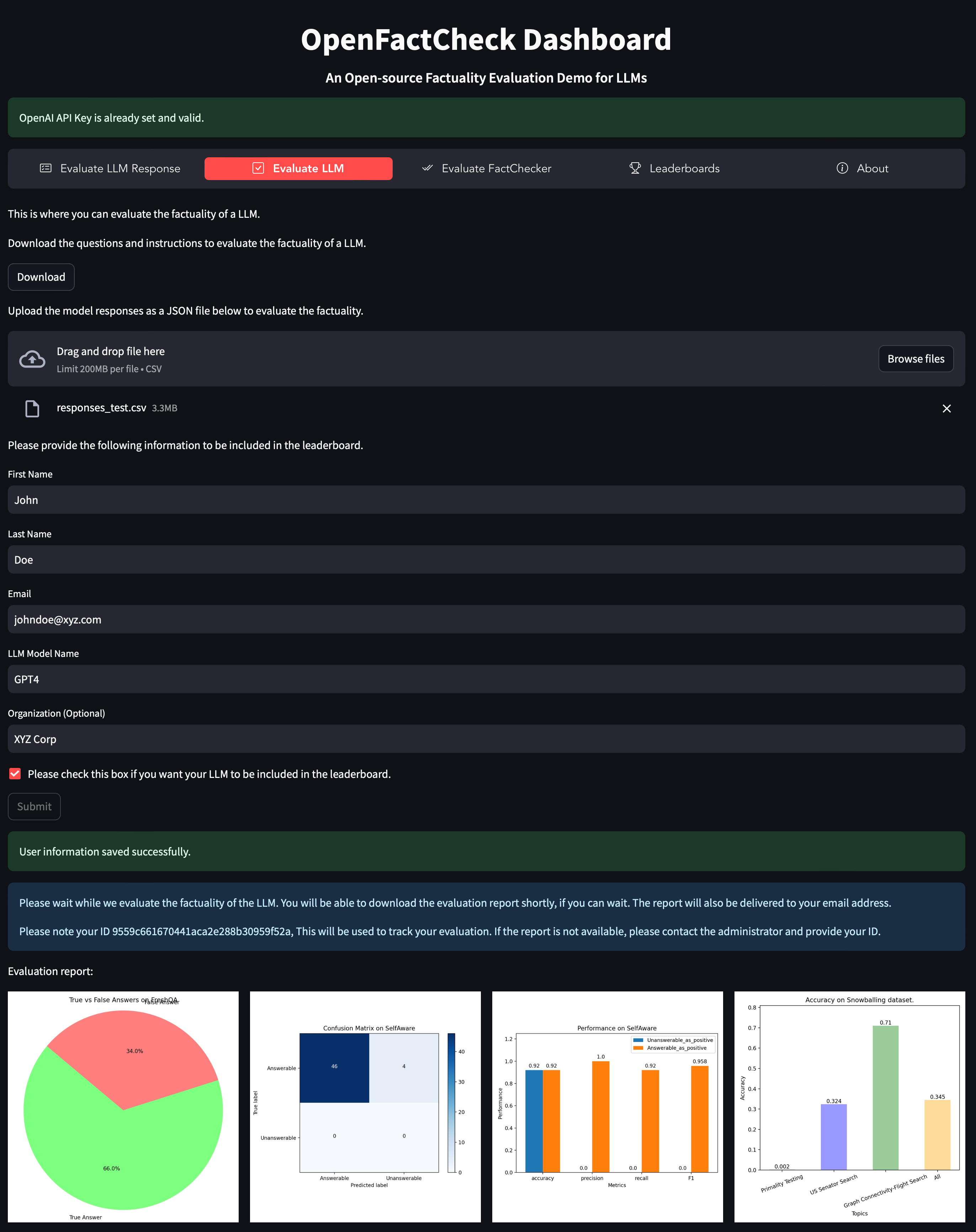}
    \caption{The interface of the LLM Factuality Evaluation page. A random evaluation result is shown for demonstration purposes.}
    \label{fig:page2}
\end{figure}

 \figref{fig:page2} corresponds to the module of \llmeval in \S\ref{sec:llmeval}. Users first download our predefined question set and then upload their model responses, the system forwards them to background tasks, using \llmeval for evaluation. Afterwards, a comprehensive report is generated and emailed to the user, notifying them of the availability of the report's PDF for download. Moreover, if users consent to publish the evaluation results, we display them on the corresponding leaderboard page. 

\begin{figure*}[ht!]
    \centering
    \includegraphics[scale=0.33]{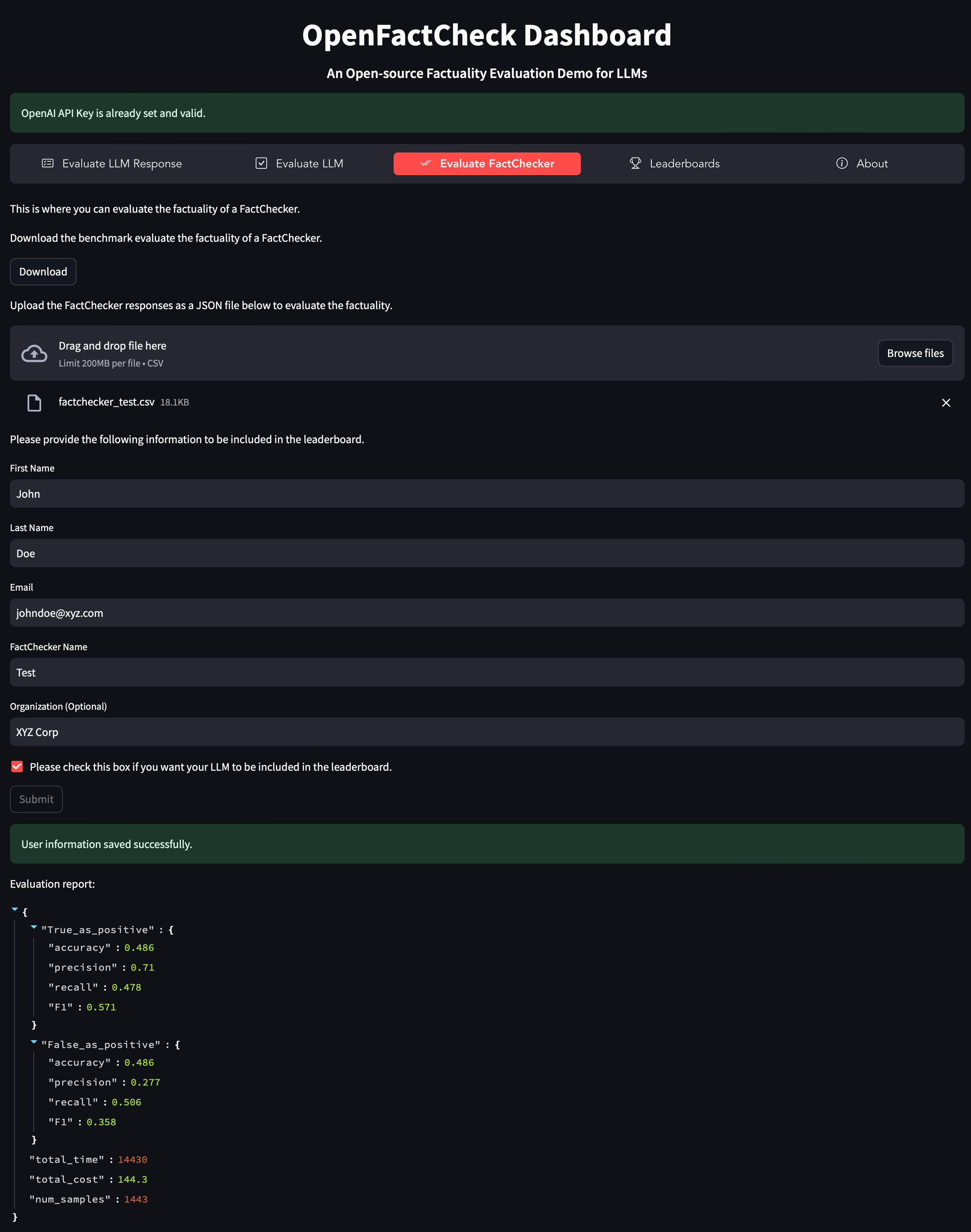}
    \caption{The interface of the Automatic Fact-checker Evaluation page. A random factchecker evaluation result is shown for demonstration purposes.}
    \label{fig:page3}
\end{figure*}

\textit{\checkereval} page in \figref{fig:page3} evaluates the performance of fact-checking systems.
Users can download claims or documents to be checked from this page, and then use their fact-checking system to predict factuality. The results including True/False, time, and USD costs are subsequently uploaded. We evaluate the submitted fact-checker results based on the ground truth labels of the human-annotated datasets, we rank and display them on the leaderboard.

\begin{figure}[ht!]
    \centering
    \includegraphics[scale=0.33]{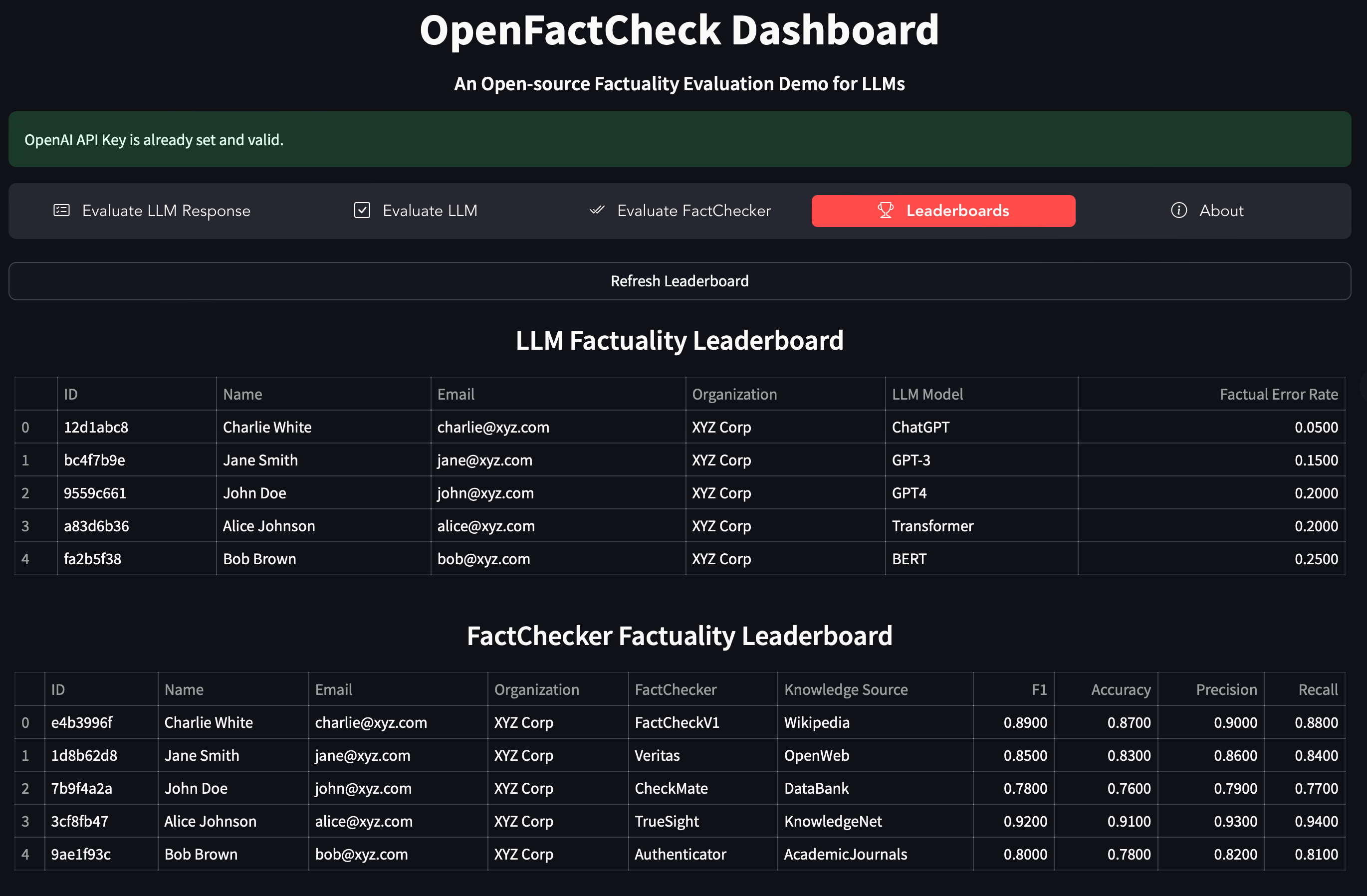}
    \caption{The interface of the Leaderboard page. Random data is shown for demonstration purposes.}
    \label{fig:page4}
\end{figure}

\textit{Leaderboard} page in \figref{fig:page4} is maintained for both the LLM factuality evaluation and the automatic fact-checking system evaluation. This leaderboard is updated in real time, allowing users to track their performance and to compare it to others. The leaderboard is accessible from the main page, providing a comprehensive overview of the system's performance. 

\end{document}